\documentclass[11pt]{article}

\usepackage[final]{acl}
\usepackage{times}          % Times Roman font (alternatives: txfonts or newtx)
\usepackage{latexsym}

\usepackage[T1]{fontenc}
\usepackage[utf8]{inputenc}

\usepackage{microtype}

\usepackage{inconsolata}
\usepackage{graphicx}
\usepackage{amsmath}
\usepackage{amssymb}
\usepackage{booktabs}
\usepackage{multirow}

\usepackage[ruled,vlined]{algorithm2e}

\title{PERSA: Reinforcement Learning for Professor-Style Personalized Feedback with LLMs }

\author{Ravi Ranjan*\\
  Florida International\\
  University (FIU), \\
  Miami, USA \\
  {\tt rkuma031@fiu.edu} \\\And
  Utkarsh Grover\\
  University of South\\
  Florida (USF), \\
  Tampa, USA \\
  {\tt utkarshgrover@usf.edu} \\\And
  Xiaomin Lin\\
  University of South\\
  Florida (USF), \\
  Tampa, USA \\
  {\tt xlin2@usf.edu} \\\And
  Agoritsa Polyzou\thanks{Corresponding authors. \\Accepted to the \textbf{ACL-2026}, BEA. \\To appear in ACL Proceedings.} \\
  Florida International\\
  University (FIU), \\
  Miami, USA \\
  {\tt apolyzou@fiu.edu} \\}

\usepackage[ruled,vlined]{algorithm2e}

\usepackage{amsmath}
\usepackage{amssymb}
\usepackage{booktabs,multirow}

\begin{document}

\maketitle
\vspace{-0.2cm}
\begin{abstract}
\vspace{-0.1cm}
Large language models (LLMs) can provide automated feedback in educational settings, but aligning an LLM’s \textbf{style} with a specific instructor’s tone while maintaining diagnostic correctness remains challenging. We ask: \textit{how can we update an LLM for automated feedback generation to align with a target instructor’s style without sacrificing core knowledge?} We study how Reinforcement Learning from Human Feedback (RLHF) can adapt a transformer-based LLM to generate programming feedback that matches a professor’s grading voice. We introduce \textbf{PERSA}, an RLHF pipeline that combines supervised fine-tuning on professor demonstrations, reward modeling from pairwise preferences, and Proximal Policy Optimization (PPO), while deliberately constraining learning to \textbf{style-bearing components}.
Motivated by analyses of transformer internals, PERSA applies parameter efficient fine-tuning. It updates only the \emph{top} transformer blocks and their feed-forward projections, minimizing global parameter drift while increasing stylistic controllability. We evaluate our proposed approach on three code-feedback benchmarks (APPS, PyFiXV, and CodeReviewQA) using complementary metrics for style alignment and fidelity. Across both Llama-3 and Gemma-2 backbones, PERSA delivers the strongest professor-style transfer while retaining correctness; for example on APPS, it boosts Style Alignment Score (SAC) to 96.2\% (from 34.8\% for Base) with Correctness Accuracy (CA) up to 100\% on Llama-3, and Gemma-2. Overall, PERSA offers a practical route to personalized educational feedback by aligning both \emph{what it says} (content correctness) and, crucially, \emph{how it says it} (instructor-like tone and structure).
\end{abstract}

\section{Introduction}
\label{sec:intro}
\vspace{-0.2cm}
Large language models (LLMs) are increasingly deployed in real-world systems, powering applications ranging from conversational assistants to code-generation tools. Major organizations such as OpenAI, Google, and Microsoft have emphasized alignment and prompt engineering to ensure these systems respond in useful and safe ways. For example, models such as ChatGPT and GPT-4 are refined through reinforcement learning from human feedback (RLHF) to better follow instructions and safety preferences \cite{ouyang2022training,chaudhari2024rlhf}, while instruction-tuning pipelines are widely used to steer tone and response style in user-facing systems \cite{wang2023aligning,bhattacharya2024demystifying}. Although these efforts have substantially improved general helpfulness and harmlessness, far less attention has been paid to \emph{personalized stylistic alignment}, where a model must communicate in a manner consistent with a particular human expert. This gap is especially important in education.

\begin{figure}[t!]
\vspace{-0.6cm}
\centering
\includegraphics[width=\linewidth]{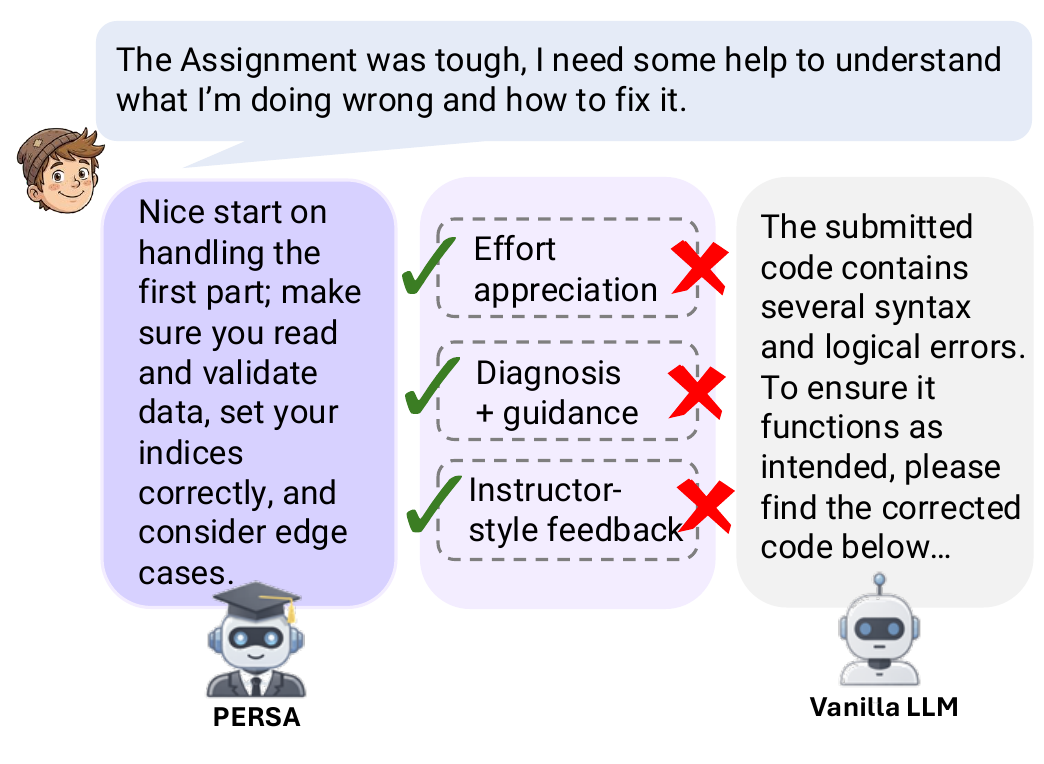}
\vspace{-0.4cm}
\caption{Illustration of how PERSA transforms generic LLM feedback into instructor-style, constructive, and actionable guidance compared to a vanilla LLM.}
\label{fig:intro}
\vspace{-0.6cm}
\end{figure}

One promising use of LLMs in education is automated feedback generation. Prior work suggests that LLM-based tutors can provide hints and critiques in domains such as programming education \cite{razafinirina2024pedagogical}. However, the effectiveness of such feedback depends not only on correctness, but also on \emph{how} it is delivered. Research in the learning sciences shows that tone, phrasing, and instructional persona strongly affect how students interpret and respond to feedback \cite{demszky2024can,loughran2019pedagogical}. Instructors often develop a recognizable style that balances encouragement with precision, emphasizing issues such as readability, correctness, and edge-case reasoning \cite{Felder2002}. Feedback that is technically correct but overly generic or terse may therefore be perceived as less trustworthy, less supportive, and less pedagogically useful \cite{carless2012trust,zhang2022fostering}. Aligning AI-generated feedback with an instructor’s authentic voice could make automated tutoring more acceptable and educationally effective.

In this work, we ask: \emph{Can a pre-trained LLM be adapted to emulate a specific instructor’s feedback style while preserving its problem-solving ability?} Existing instruction-tuned models are typically optimized for broad helpfulness and politeness, but alignment to an individual educator’s style remains underexplored. To address this, we propose \textbf{Professor-Style Reinforcement-based Style Adaptation (PERSA)}, a method that uses RLHF to align model outputs with the feedback style of a target professor. Using a dataset of instructor-authored feedback on student work, we train a reward model that captures stylistic preferences and optimize the policy for \emph{stylistic fidelity} rather than generic user satisfaction. As shown in Figure~\ref{fig:intro}, whereas a vanilla LLM might generate a brief suggestion such as ``Check your input handling,'’ professor-style feedback is typically more explanatory, supportive, and actionable. PERSA narrows this gap by producing feedback that more closely reflects the instructor’s voice, including encouragement, diagnosis, corrective guidance, and verification-oriented follow-up. In doing so, it supports seamless personalization of AI tutors for specific courses and teaching contexts.

A central technical contribution of PERSA is a \emph{style-targeted, parameter-efficient fine-tuning strategy}. Rather than updating all model weights, which is computationally costly and risks degrading core capabilities, we focus adaptation on components most relevant to stylistic expression. Prior work suggests that many style-related properties are concentrated in higher transformer layers \cite{lai2024style,koto2022neuron}. Guided by this observation, we use low-rank adaptation \cite{hu2021lora} to fine-tune only the final layers of a 3B-parameter LLM. This selective update strategy preserves the model’s underlying problem-solving competence while shifting its communicative ``voice,'' reducing the catastrophic forgetting that can arise from indiscriminate fine-tuning \cite{Luo2023}. As a result, PERSA captures subtle instructor-specific traits, such as level of detail and encouraging tone, without compromising content accuracy.

We evaluate PERSA on a programming feedback generation task against strong baselines, including the base pre-trained model, supervised fine-tuning on the same instructor data, and an InstructGPT-style RLHF model optimized for general feedback quality. PERSA achieves the best overall balance between style and correctness. Its outputs show strong stylistic alignment with professor feedback, on  Gemma-2 model with a Style Alignment Score approaching 0.99, and BLEU overlap above 98\%, substantially exceeding competing methods \cite{papineni2002bleu,hu2023style}. At the same time, correctness remains at 100\%, and feedback retains a consistently polite and constructive tone as measured by a politeness model \cite{danescu2013computational}. Human evaluators also strongly prefer PERSA, often describing its responses as reading ``like they came from the professor.'' These findings suggest that combining RLHF with targeted layer adaptation is effective for capturing nuanced pedagogical style beyond what standard fine-tuning or generic alignment methods can achieve.

In summary, this paper makes three contributions: \textbf{(1)} we introduce \textbf{PERSA}, an RLHF-based framework for aligning LLM feedback with a specific instructor’s pedagogical style; \textbf{(2)} we develop a \emph{layer-selective} LoRA-based adaptation strategy that achieves stylistic transformation with minimal disruption to the model’s original capabilities; and \textbf{(3)} we provide comprehensive evaluation, including automated style metrics and a human study, showing that PERSA produces feedback that is both pedagogically authentic and factually correct. Overall, this work advances the development of AI tutors that can not only provide accurate guidance but also communicate with the tone and effectiveness of a human instructor.

\vspace{-0.1cm}
\section{Related Work}
\label{sec:related}
\vspace{-0.1cm}

\textbf{Educational Feedback and Pedagogical Style.}
Prior work in educational technology shows that the \emph{manner} of feedback delivery, including tone, politeness, and personalization, substantially affects learner engagement and outcomes \cite{han2023llm, sonlu2024effects}. Recent studies on LLMs in education similarly emphasize that encouragement, specificity, and perceived instructor authenticity influence student trust and uptake \cite{razafinirina2024pedagogical, wu2024analyzing}. While this connects to broader alignment goals in dialogue systems \cite{alsafari2024towards, wang2023aligning}, our focus is narrower: \emph{pedagogical alignment}, i.e., adapting model outputs to an instructor’s communicative style rather than optimizing only for generic helpfulness \cite{bhattacharya2024demystifying,matarazzo2025survey}.

\textbf{Automated Feedback Generation.}
Automated programming feedback has long been studied through rule-based, analysis-driven, and repair-based methods. Keuning et al.\ provide a foundational survey of these approaches and their evaluation limitations \cite{Keuning2018}, while \emph{Pedal} demonstrates practical code-aware feedback generation at scale \cite{Gusukuma2020}. More recently, LLMs have shown promise but also reliability concerns: GPT-4 can generate readable feedback yet still requires validation \cite{Dai2024}; MOOC-scale studies report that LLM feedback helps detect errors but is often inaccurate without test-based grounding \cite{Gabbay2024}; and performance drops further in specialized settings such as concurrency debugging \cite{EstevezAyres2024}. Other efforts explore diversified repair generation \cite{Choi2025} and responsible deployment considerations such as inclusivity and hallucination risk \cite{lindsay2025responsible}. Collectively, these works highlight that existing systems still under-address instructor-consistent tone and specificity.

\textbf{RLHF for Language Model Alignment.}
RLHF has become a standard paradigm for aligning LLMs to human preferences \cite{ouyang2022training, xie2024sorry}. InstructGPT established the now-common pipeline of supervised fine-tuning, reward modeling, and PPO-based policy optimization \cite{ouyang2022training}, and subsequent work has extended RLHF to capture subtler properties such as tone, framing, and politeness \cite{sinha2025instruction, kirk2023understanding,dong2024rlhf,yan2024reward}. This is particularly relevant in education, where instructor judgments define high-quality feedback. However, full-parameter RLHF is computationally expensive and can unintentionally alter broader model capabilities \cite{chaudhari2024rlhf}. Parameter-efficient alternatives, including LoRA-augmented RLHF and related methods, reduce this cost while preserving alignment quality \cite{sidahmed2024perl,hong2024orpo,ethayarajh2024kto,hu2021lora,dettmers2023qlora,wu2024beta}. Importantly, prior analyses suggest that RLHF-induced changes are concentrated in upper transformer layers, especially FFN components \cite{tigges2023exploratory}, motivating our layer-selective adaptation strategy.

\textbf{Style Adaptation in Language Models.}
Style control in LLMs has been explored through prompting, style tokens, and fine-tuning, with recent work moving toward neuron- and layer-level analysis. Lai et al.\ show that stylistic attributes are disproportionately localized in higher transformer layers and FFN submodules, enabling targeted control via style-specific neurons \cite{lai2024style,shi2024understanding}. Related studies further suggest that selectively tuning or freezing layers can outperform full-model adaptation for certain alignment objectives \cite{shi2024understanding,wang2023survey}. Parameter-efficient methods such as LoRA and QLoRA make such targeted adaptation practical, though quantized tuning may introduce accuracy trade-offs in some settings \cite{dettmers2023qlora, hu2021lora,ozdemir2023quick}. In contrast to prior style-control work, PERSA applies these insights to \emph{educational feedback}, aligning LLM outputs to an individual instructor’s pedagogical voice while preserving correctness.
\vspace{-0.1cm}
\section{Preliminaries}
\label{sec:prelim}
\vspace{-0.1cm}
\textbf{Transformer LLMs as policies.}
We consider a decoder-only transformer language model as a stochastic policy that generates a feedback sequence $y=(y_1,\ldots,y_T)$ token-by-token
conditioned on a prompt $x$ (e.g., problem statement + student solution).

\textbf{Data: demonstrations and preferences.}
We assume (i) an instructor demonstration set
$\mathcal{D}_{\text{SFT}}=\{(x_i, y_i^\star)\}_{i=1}^{N}$ with professor-written feedback
$y_i^\star$, and (ii) a preference set
$\mathcal{D}_{\text{pref}}=\{(x_j, y_j^{(w)}, y_j^{(l)})\}_{j=1}^{M}$, where $y^{(w)}$ is
preferred to $y^{(l)}$ under instructor style (and typically correctness). Preference
data is used to learn a reward model for RLHF \cite{ouyang2022training}.

\textbf{Supervised fine-tuning (SFT).}
SFT initializes the policy by minimizing the teacher-forcing negative log-likelihood:
\begin{equation}
\mathcal{L}_{\text{SFT}}(\theta)
= - \mathbb{E}_{(x,y^\star)\sim \mathcal{D}_{\text{SFT}}}
\Big[\sum_{t=1}^{|y^\star|}
\log \pi_{\theta}(y_t^\star \mid x, y^\star_{<t})\Big].
\label{eq:sft_prelim}
\end{equation}
Here $y^\star_{<t}$ is the the gold prefix tokens before step $t$.

\textbf{Reward modeling from pairwise comparisons.}
RLHF learns a scalar reward function $r_{\phi}(x,y)\in\mathbb{R}$ such that preferred
responses receive higher reward. Using the Bradley--Terry model, the probability that
$y^{(w)}$ is preferred over $y^{(l)}$ is
\begin{equation}
\Pr(y^{(w)} \succ y^{(l)} \mid x)
= \sigma\!\big(r_{\phi}(x,y^{(w)}) - r_{\phi}(x,y^{(l)})\big),
\label{eq:bt_prelim}
\end{equation}
yielding the standard pairwise logistic loss, and $\sigma$ is the sigmoid
function~\cite{ouyang2022training}.

\textbf{Policy optimization with PPO and KL control.}
Given $r_{\phi}$, policy optimization seeks high reward while constraining drift from a
reference policy $\pi_{\text{ref}}$ (often the post-SFT model) via a KL penalty.

We optimize this objective with Proximal Policy Optimization (PPO) \cite{schulman2017proximal}.
Let $s_t=(x,y_{<t})$ be the state (prefix context), $a_t=y_t$ the action (next token),
and $\rho_t(\theta)=\pi_{\theta}(a_t|s_t)/\pi_{\theta_{\text{old}}}(a_t|s_t)$.

\textbf{Parameter-efficient adaptation with LoRA.}
To preserve base model knowledge and reduce computation, PERSA adopts Low-Rank Adaptation
(LoRA) \cite{hu2021lora}. For a frozen weight matrix $W\in\mathbb{R}^{d\times k}$, LoRA
learns a low-rank update $\Delta W$ with $B\in\mathbb{R}^{d\times r}$,
$A\in\mathbb{R}^{r\times k}$ and $r\ll \min(d,k)$:
\begin{equation}
W' = W + \Delta W,\quad \text{where} ~~\Delta W = BA.
\label{eq:lora_prelim}
\end{equation}
We denote the trainable adapter parameters by $\theta_{\text{LoRA}}$ and treat the
remaining parameters as frozen. Parameter-efficient RLHF variants (e.g., PERL) show this
can substantially reduce training cost while maintaining alignment quality \cite{sidahmed2024perl}.

\begin{figure*}[bt]
    \centering
    \includegraphics[width=0.95\linewidth]{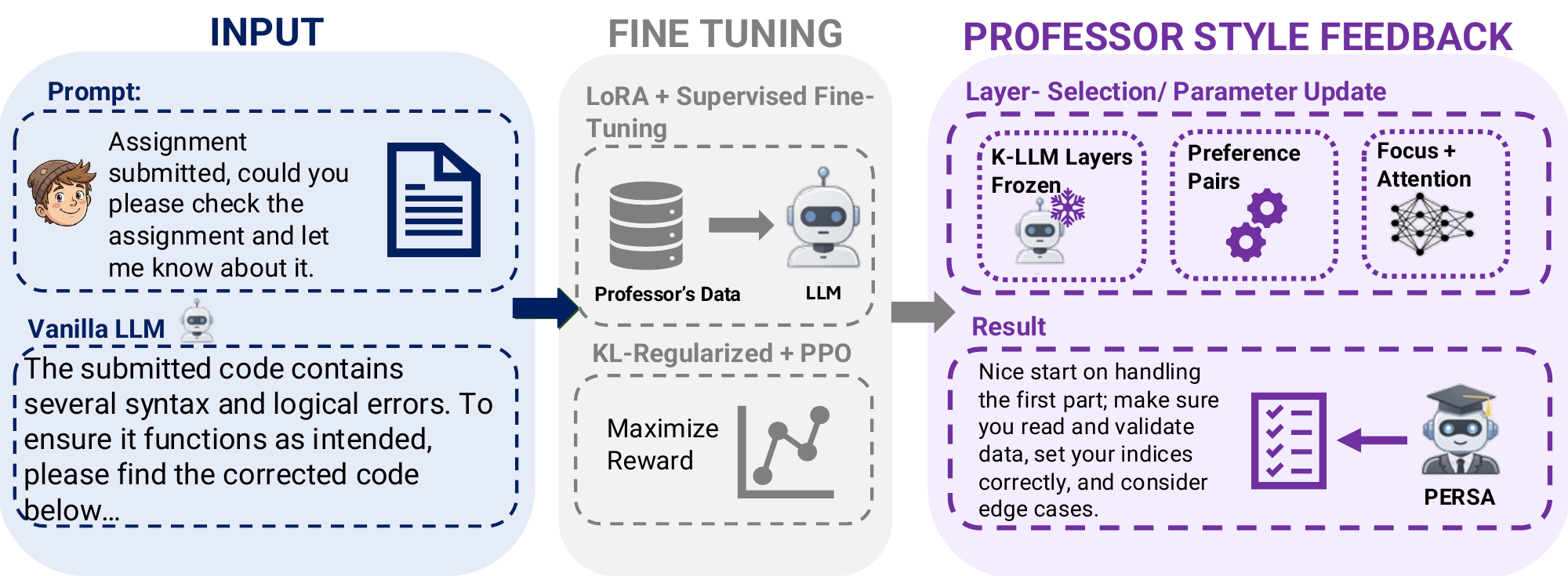} 
    \caption{PERSA pipeline for professor-style feedback alignment via layer-selective RLHF.}
    \label{fig:method}
    \vspace{-0.4cm}
\end{figure*}

\textbf{Layer-selective tuning.}
Finally, we use the notion of \emph{top-layer} tuning: only adapters inserted in the last
$L$ transformer blocks are updated. This choice is consistent with evidence that higher
layers and FFN submodules often contain high-level attributes and alignment-relevant
representations \cite{geva2021transformer, lai2024style}.

\section{Methodology}
\label{sec:method}
We propose \textbf{PERSA} (Professor-Style Reinforcement-based Style Adaptation), a parameter efficient RLHF framework that aligns an LLM's \emph{feedback style} to a target instructor while preserving the model's underlying problem-solving and language competence. PERSA instantiates the standard RLHF recipe, \emph{supervised fine-tuning} $\rightarrow$ \emph{reward modeling} $\rightarrow$ \emph{policy optimization}, popularized by InstructGPT~\cite{ouyang2022training}, but crucially restricts learning to \emph{style-bearing} components in the \emph{upper} transformer layers using low-rank adapters~\cite{hu2021lora,sidahmed2024perl}. This design is motivated by growing evidence that higher layers and FFN submodules disproportionately encode high-level attributes such as discourse patterns, tone, and stylistic preferences~\cite{geva2021transformer,lai2024style}. Figure~\ref{fig:method} illustrates how PERSA transforms a generic prompt into instructor-style feedback by combining LoRA-based supervised fine-tuning with KL-regularized PPO (trained on professor data).

\textbf{Problem Setup and Notation.}
Let $\pi_{\theta}$ denote a pretrained decoder-only transformer LLM (e.g., LLaMA/Gemma/GPT-style), parameterized by $\theta$. Each training instance consists of a prompt $x$ and a target feedback $y^\star$ written by the professor. The prompt $x$ contains a programming problem statement and a student submission. The goal is to learn a policy $\pi_{\theta}$ that produces feedback $y$ that is simultaneously:
(i) \emph{diagnostically correct} (identifies correctness/errors and provides actionable guidance) and
(ii) \emph{stylistically aligned} with the professor (tone, structure, politeness, specificity).

We assume access to (a) demonstration data
$\mathcal{D}_{\text{SFT}}= \{(x_i,y_i^\star)\}_{i=1}^{N}$,
and (b) preference comparisons
$\mathcal{D}_{\text{pref}}= \{(x_i,y_i^{(w)},y_i^{(l)})\}_{i=1}^{M}$,
where $y^{(w)}$ is preferred over $y^{(l)}$ under instructor style (and typically correctness).

\textbf{Parameter-Efficient, Layer-Selective Updates.}
\label{subsec:lora}
Rather than updating all parameters, we freeze the base model weights and learn low-rank adapter updates (LoRA) only in the \emph{top} $L$ transformer blocks (e.g., $L=4$), focusing on attention projections and Feed-Forward Network projections. For a weight matrix $W\in\mathbb{R}^{d\times k}$ in a selected module, LoRA parameterizes, as detailed mentioned in equation~\ref{eq:lora_prelim}.
This yields a small trainable parameter set, $\theta_{\text{LoRA}}$ (on the order of $\sim 10^7$ parameters in our setting), while leaving the remaining parameters fixed, improving stability and reducing risk of catastrophic drift~\cite{sidahmed2024perl}. Our layer-selection choice is supported by analyses indicating that alignment- and style-related changes concentrate in later layers, often in the penultimate FFNs~\cite{geva2021transformer,liang2024can}.

\subsection{Supervised Fine-Tuning (SFT)}

We first perform instruction tuning to initialize a policy that produces professor-like feedback under maximum likelihood. The SFT objective is the standard auto-regressive negative log-likelihood, defined in equation~\ref{eq:sft_prelim}.
where only $\theta_{\text{LoRA}}$ is updated and all other parameters are frozen. This stage transfers the professor's feedback structure (e.g., diagnosis $\rightarrow$ correction $\rightarrow$ verification) and baseline tone into the model, but may not optimally capture preference-sensitive nuances (e.g., strictness, encouragement, emphasis on edge cases).

\subsection{RLHF for Professor-Style Alignment}
\label{subsec:rlhf}
SFT imitates demonstrations but does not explicitly optimize for \emph{human preference}. We therefore apply RLHF, consisting of reward modeling followed by PPO-based policy optimization, following established alignment practice~\cite{ouyang2022training}.

\subsubsection{Reward Modeling from Pairwise Preferences}
We train a reward model $r_{\phi}(x,y)\in\mathbb{R}$ that scores candidate feedback $y$ for prompt $x$. The reward model is a transformer initialized from a strong LM backbone with an added scalar head, trained on pairwise comparisons using the Bradley Terry formulation:
\begin{equation}
\Pr\!\left(y^{(w)} \succ y^{(l)} \mid x\right)
=
\sigma\!\left(r_{\phi}(x,y^{(w)})-r_{\phi}(x,y^{(l)})\right),
\label{eq:bt}
\end{equation}
where $\sigma$ is the sigmoid function. Minimizing the negative log-likelihood yields:
\begin{equation}
\begin{split}
\mathcal{L}_{\text{RM}}(\phi)
=
-\mathbb{E}_{(x,y^{(w)},y^{(l)})\sim\mathcal{D}_{\text{pref}}} \\
 \left[\log \sigma\!\left(r_{\phi}(x,y^{(w)})-r_{\phi}(x,y^{(l)})\right)\right].
\label{eq:rm}
\end{split}
\end{equation}
Because preferred responses are professor-authored (or instructor-preferred), $r_{\phi}$ learns to jointly reflect correctness and stylistic fidelity, mirroring the RLHF setup in InstructGPT but with an education-specific preference signal~\cite{ouyang2022training}.

\subsubsection{Policy Optimization with PPO and KL Control}
Given $r_{\phi}$, we optimize the policy to maximize expected reward while constraining deviation from a reference policy (the SFT model) via a KL penalty. Let $\pi_{\theta}$ be the current policy and $\pi_{\text{ref}}$ a frozen copy of the SFT policy. We optimize:
\begin{equation}
\label{eq:rl_obj}
\begin{split}
\max_{\theta_{\text{LoRA}}} \quad
& \mathbb{E}_{x\sim\mathcal{D},\, y\sim\pi_{\theta}(\cdot|x)}
\Big[
r_{\phi}(x,y)
\\
&\qquad
-\beta\,\mathrm{KL}\!\left(
\pi_{\theta}(\cdot|x)
\,\|\, 
\pi_{\text{ref}}(\cdot|x)
\right)
\Big]
\end{split}
\end{equation}
where $\beta>0$ controls the strength of KL regularization, a key stabilizer in RLHF~\cite{ouyang2022training}.

We solve Eq.~\eqref{eq:rl_obj} with PPO~\cite{schulman2017proximal}. For token-level actions $a_t$ under states $s_t$ (prefix context), we define the importance ratio
$\rho_t(\theta)=\pi_{\theta}(a_t|s_t)/\pi_{\theta_{\text{old}}}(a_t|s_t)$ and advantage estimate $\hat{A}_t$.
The clipped PPO objective is:
 \begin{equation}
\begin{split}
\mathcal{L}_{\text{PPO}}(\theta_{\text{LoRA}}) = -\mathbb{E}_t \bigl[ & \min\!\bigl( \rho_t(\theta)\,\hat{A}_t, \\
& \mathrm{clip}(\rho_t(\theta),1-\epsilon,1+\epsilon)\,\hat{A}_t \bigr) \bigr],
\end{split}
\label{eq:ppo}
\end{equation}
with clipping threshold $\epsilon$.
We additionally include the KL control term (computed against $\pi_{\text{ref}}$) in the per-trajectory reward or as an auxiliary penalty, consistent with the constrained-RL perspective in RLHF~\cite{ouyang2022training}. As in SFT, we update only $\theta_{\text{LoRA}}$ in the selected upper layers, yielding a parameter-efficient instantiation of RLHF akin to PERL~\cite{sidahmed2024perl}. This selective optimization reduces compute and helps preserve the pretrained model’s broad capabilities.

\textbf{Summary.} PERSA proceeds as follows:
(i) attach LoRA adapters to the top $L$ layers (\S\ref{subsec:lora}),
(ii) run SFT on $\mathcal{D}_{\text{SFT}}$ minimizing Eq.~\eqref{eq:sft_prelim},
(iii) train a reward model on $\mathcal{D}_{\text{pref}}$ minimizing Eq.~\eqref{eq:rm},
and (iv) run PPO to optimize Eq.~\eqref{eq:ppo} under the KL-regularized objective Eq.~\eqref{eq:rl_obj}. The related steps are outlined in Alg.~\ref{alg:persa_brief}, Appendix~\ref{app:pseudo}.
This pipeline is architecture-agnostic for modern decoder-only LLMs and is especially suitable for educational personalization, where the goal is to preserve task competence while adapting communicative style.

\section{Experimental Setup}
\label{sec:experiments}
\vspace{-0.2cm}
\subsection{Datasets}
\label{subsec:datasets}
\vspace{-0.2cm}
We evaluate PERSA on three code-feedback datasets: one instructor-authored APPS-style professor-feedback dataset and two public code-feedback benchmarks, PyFiXV and CodeReviewQA. We construct a standard 70/10/20 train/validation/test split with disjoint programs for \textsc{PERSA} training and evaluation.

\textbf{(i) APPS Benchmark (professor-feedback dataset).}
To study realistic instructional personalization, we use a course-specific dataset of 200 instances, each consisting of an algorithmic programming problem, a student solution (either correct or deliberately perturbed), and \emph{professor-written feedback} as the target response. The prompts are derived from APPS-style problems, while the feedback is authored by the instructor for the target course~\cite{hendrycks2021measuring}. 

\textbf{(ii) Codeforces Syntax-Error Feedback (PyFiXV).}
We use the Codeforces subset of PyFiXV, which contains 240 Python programs with syntax errors (token length $\leq 500$), collected from Codeforces and paired with expert-written fixes and explanations~\cite{phung2023generating}.

\textbf{(iii) \texttt{Tomo-Melb/CodeReviewQA}.}
CodeReviewQA comprises 900 manually curated code-review instances spanning 9 programming languages. Each example includes a pre-review snippet (\texttt{old}), a reviewer comment (\texttt{review}), and the revised code (\texttt{new}). We cast this benchmark as a text-to-text task, mapping $(x=\{\texttt{old},\texttt{review}\})$ to $y=\texttt{new}$ \cite{lin2025codereviewqa}.

The appendix Table~\ref{tab:leakage_controls} summarizes data splits \& leakage controls.
\vspace{-0.1cm}
\subsection{Model Evaluation Frameworks}
\label{subsec:models}
\vspace{-0.1cm}
We evaluate \textsc{PERSA} on two lightweight, open-weight instruction-tuned backbones, \texttt{google/gemma-2-2b-it} and \texttt{meta-llama/Llama-3.2-3B-Instruct}, within a model-agnostic RLHF framework comprising supervised fine-tuning, reward modeling, and PPO-based policy optimization with KL regularization. We compare against a diverse set of baselines, including the untuned \textbf{Base Model}, \textbf{SFT}, \textbf{InstructGPT-style RLHF}, and three offline preference-optimization methods, namely \textbf{DPO}, \textbf{ORPO}, and \textbf{KTO}. This setup enables a controlled comparison of stylistic alignment, correctness preservation, and adaptation efficiency across both model families under a shared evaluation protocol. Full details of the backbones, training pipeline, baseline configurations, and evaluation settings are provided in Appendix~\ref{app:sec-model-eval}.

\subsection{Evaluation Metrics:}
\label{subsec:evaluation-metric}
\vspace{-0.1cm}

We evaluate feedback generation along two complementary dimensions: \emph{stylistic alignment} and \emph{diagnostic fidelity}. To quantify how closely the generated feedback matches the instructor’s communicative style, we use \textbf{Style Alignment Score (SAC)}~\cite{hu2023style}, which estimates the probability that a response reflects the professor’s stylistic characteristics. We further measure tone similarity using \textbf{Average Politeness Closeness (APC)}~\cite{danescu2013computational}, which evaluates how closely the politeness level of the generated feedback aligns with the professor reference. Finally, we report \textbf{BLEU-4}~\cite{papineni2002bleu} to capture lexical similarity by measuring n-gram overlap between generated feedback and the professor’s reference response.
All together it measure stylistic fidelity to the professor’s feedback in terms of overall style alignment, politeness closeness, and lexical similarity, respectively, while Correctness Accuracy \textbf{(CA)}~\cite{shen2023large} assesses whether the generated feedback makes the correct diagnostic judgment about the student solution. In addition, Preference Win Rate \textbf{(PWR)}~\cite{ouyang2022training} measures preference-based superiority relative to competing methods, reflecting the alignment objective used in RLHF. Full metric definitions, formulations, and implementation details are provided in Appendix~\ref{app:eval-metric}. For configuration see Appendix~\ref{app:exp-config}.

\begin{table*}[t]
\centering
\scriptsize
\setlength{\tabcolsep}{2.0pt}
\renewcommand{\arraystretch}{0.90}
% \vspace{-2mm}
\caption{Comparison of Llama-3 vs. Gemma-2 models across three datasets (APPS, PyFiXV, and CodeReviewQA). Higher values indicate better performance; all metrics are reported in percentages.}
\label{tab:main-result}
\resizebox{\textwidth}{!}{%
\begin{tabular}{ll|ccccc|ccccc}
\toprule
\multirow{2}{*}{Dataset} & \multirow{2}{*}{Method} &
\multicolumn{5}{c|}{\textbf{Llama-3 (\%)}} &
\multicolumn{5}{c}{\textbf{Gemma-2 (\%)}} \\
\cmidrule(lr){3-7}\cmidrule(lr){8-12}
& & SAC & APC & BLEU-4 & CA & PWR & SAC & APC & BLEU-4 & CA & PWR \\
\midrule
\multirow{7}{*}{APPS}
& Base & 34.8 & 85.0 & 6.4 & 98.2 & -- & 20.0 & 90.0 & 2.0  & 98.0 & -- \\
& SFT & 82.0 & 86.0 & 80.0 & 100  & 86.2 & 85.0 & 95.0 & 70.0 & 100  & 90.0 \\
& InstructGPT/RLHF & 84.0 & 86.4 & 80.0 & 100  & 88.0 & 88.0 & 95.0 & 80.0 & 100  & 95.0 \\
& DPO & 94.0 & 87.8 & 94.2 & 100  & 89.8 & 86.0 & 95.0 & 78.0 & 100  & 94.0 \\
& ORPO & 95.6 & 89.0 & 95.0 & 100  & \textbf{90.2} & 89.0 & 95.0 & 80.0 & 100  & 98.0 \\
& KTO & 95.0 & 87.2 & 94.6 & 100  & 90.0 & 90.0 & 95.0 & 81.0 & 100  & 97.2 \\
& \textbf{PERSA} & \textbf{96.2} & \textbf{92.1} & \textbf{95.8} & \textbf{100} & 90.1
              & \textbf{99.0} & \textbf{95.0} & \textbf{98.0} & \textbf{100} & \textbf{98.0} \\
\midrule
\multirow{7}{*}{PyFiXV}
& Base & 30.5 & 84.0 & 8.0 & 97.5 & -- & 20.0 & 90.0 & 5.0  & 96.0 & -- \\
& SFT & 79.0 & 85.5 & 76.0 & 99.5 & 84.5 & 80.0 & 94.0 & 60.0 & 100  & 85.0 \\
& InstructGPT/RLHF & 81.0 & 86.0 & 77.5 & 99.7 & 86.0 & 82.0 & 95.0 & 70.0 & 100  & 90.0 \\
& DPO & 92.0 & 87.0 & 92.5 & 99.8 & 88.0 & 81.0 & 95.0 & 68.0 & 100  & 88.0 \\
& ORPO & 93.5 & 88.0 & 93.2 & 99.8 & 88.6 & 83.0 & 95.0 & 72.0 & 100  & 98.0 \\
& KTO & 93.0 & 87.0 & 93.0 & 99.8 & 88.3 & 84.0 & 95.0 & 73.0 & 100  & 96.0 \\
& \textbf{PERSA} & \textbf{94.5} & \textbf{91.0} & \textbf{94.0} & \textbf{99.9} & \textbf{89.0}
              & \textbf{98.0} & \textbf{95.0} & \textbf{98.0} & \textbf{100} & \textbf{98.2} \\
\midrule
\multirow{7}{*}{CodeReviewQA}
& Base & 28.0 & 83.5 & 8.0 & 96.8 & -- & 15.0 & 90.0 & 3.0  & 95.0 & -- \\
& SFT & 78.0 & 85.0 & 75.0 & 99.2 & 83.5 & 80.0 & 94.0 & 65.0 & 100  & 85.0 \\
& InstructGPT/RLHF & 80.0 & 85.5 & 76.5 & 99.5 & 85.0 & 84.0 & 95.0 & 75.0 & 100  & 90.0 \\
& DPO & 91.0 & 86.8 & 92.0 & 99.7 & 87.0 & 82.0 & 95.0 & 73.0 & 100  & 89.0 \\
& ORPO & 92.2 & 87.5 & 92.7 & 99.7 & 87.5 & 86.0 & 95.0 & 76.0 & 100  & 98.0 \\
& KTO & 92.0 & 86.5 & 92.5 & 99.7 & 87.3 & 87.0 & 95.0 & 78.0 & 100  & 96.0 \\
& \textbf{PERSA} & \textbf{93.8} & \textbf{90.5} & \textbf{93.5} & \textbf{99.8} & \textbf{88.0}
              & \textbf{98.0} & \textbf{95.0} & \textbf{98.0} & \textbf{100} & \textbf{98.2} \\
\bottomrule
\end{tabular}%
\vspace{-0.4cm}
}

\end{table*}

\section{Results}
\label{subsec:result}
\vspace{-0.2cm}
\subsection{Performance Comparison}
\begin{figure}[t]
\centering
\includegraphics{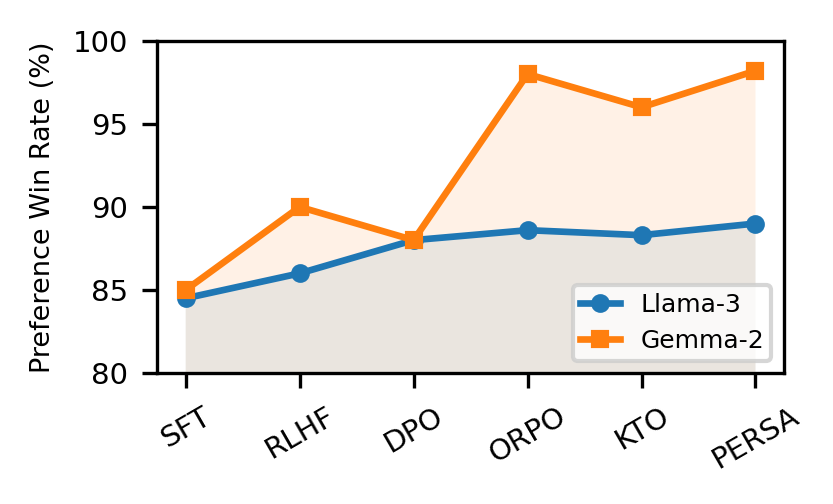}
\caption{PWR comparison across alignment methods on the PyFiXV dataset for Llama-3 and Gemma-2.}
\label{fig:pwr}
\vspace{-0.4cm}
\end{figure}

Table~\ref{tab:main-result} highlights consistent method-wise trends across all three datasets and both backbones. First, \textbf{Base} models achieve high correctness (CA$\approx$96--98\%) and moderate politeness (APC$\approx$83--90\%) but \textbf{very weak stylistic/lexical alignment} (low SAC and single-digit BLEU-4), indicating that pretraining alone does not recover professor-style feedback. Second, \textbf{SFT} provides the largest jump in alignment (e.g., SAC$\approx$78--85\%, BLEU-4$\approx$60--80\%) while keeping CA near-perfect, showing that demonstrations effectively teach the core feedback structure. Third, preference-based alignment further improves style: DPO/ORPO/KTO consistently outperform InstructGPT/RLHF in SAC and BLEU-4 with comparable CA, and yield higher win-rates (PWR), suggesting that direct preference optimization better captures subtle stylistic cues. \textbf{Figure~\ref{fig:pwr}} shows that PERSA delivers consistently stronger human preference alignment than SFT, RLHF, and DPO-style baselines on PyFiXV, with improvements that generalize across both Llama-3 and Gemma-2 backbones.
Finally, PERSA achieves the strongest overall results on both models, delivering the highest SAC and BLEU-4 and the best PWR across datasets (with CA$\approx$100\%), demonstrating robust professor-style transfer without sacrificing diagnostic correctness. 

\begin{figure}[t!]
\centering
\includegraphics{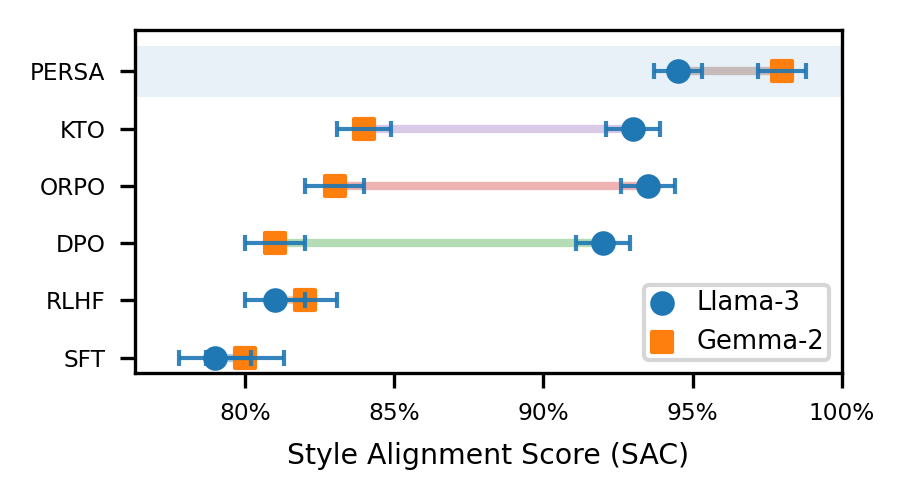}
\caption{Dumbbell plot of SAC on PyFiXV across alignment methods, comparing Llama-3 and Gemma-2; connectors indicate the cross-model gap per method (higher is better).}
\label{fig:sac}
\vspace{-0.4cm}
\end{figure}

\subsection{Qualitative Analysis}
\label{app:qual-analysis}

\begin{table}[t]
\centering
\scriptsize
\setlength{\tabcolsep}{3pt}
\renewcommand{\arraystretch}{1.08}
\caption{\small \textbf{Qualitative comparison (one prompt).} PERSA produces the most professor-like feedback: structured diagnosis $\rightarrow$ actionable fix $\rightarrow$ verification. Preference-optimization methods (ORPO/KTO/DPO) are typically more specific than generic RLHF, while Base tends to be vague.}
\label{tab:qualitative-result}
\begin{tabular}{p{0.23\columnwidth} p{0.72\columnwidth}}
\toprule
\textbf{Method} & \textbf{Qualitative feedback snippet (same prompt)} \\
\midrule
\multicolumn{2}{p{0.95\columnwidth}}{\textbf{Prompt (APPS-style).} Student computes \texttt{x=a.index(1), y=a.index(n)} before \texttt{a=list(read())}, causing a runtime error / wrong output.} \\
\midrule
Base & ``Try checking input handling and edge cases. Consider swapping min/max positions, then rerun tests.'' \\
SFT & ``You are calling \texttt{a.index()} before initializing \texttt{a}. Move \texttt{a=list(read())} above the \texttt{index} calls, then recompute \texttt{x,y}.'' \\
InstructGPT & ``Good start. The issue is order of operations: read the array first, then find indices and apply the swap. Verify with sorted/reversed cases.'' \\
DPO & ``Root cause: \texttt{a} is used before assignment. Fix by reading input into \texttt{a} first, then compute \texttt{x=a.index(1)} and \texttt{y=a.index(n)}. Add tests for \(n{=}1\) and already-sorted.'' \\
ORPO & ``Bug: \texttt{a.index()} runs before \texttt{a=list(read())}. \textbf{Initialize \texttt{a} first}, then compute \texttt{x,y} and perform one swap maximizing \(|x{-}y|\). \textbf{Re-run edge cases} (sorted, reversed, single displacement).'' \\
KTO & ``You are using \texttt{a} before it is populated. Read input into \texttt{a} first, then compute indices and swap. Avoid over-claiming; confirm correctness with boundary cases and failing tests.'' \\
\textbf{PERSA} & \textbf{``Good progress on locating 1 and \(n\). The failure arises from using \texttt{a.index()} before \texttt{a} is populated. Please first read input with \texttt{a=list(read())}; then recompute \texttt{x=a.index(1), y=a.index(n)} and apply one swap maximizing \(|x{-}y|\). Finally, rerun edge cases (already-sorted, reversed, \(n{=}1\)) to confirm.''} \\
\bottomrule
\end{tabular}
\vspace{-3mm}
\end{table}

\noindent We sample representative programming prompts from the APPS dataset and reproduce the corresponding feedback outputs. \textbf{Table~\ref{tab:qualitative-result}} mirrors the quantitative rankings in Table~\ref{tab:main-result}: methods with higher SAC/BLEU-4 (DPO/ORPO/KTO, and especially PERSA) produce feedback that is \emph{more diagnosis-specific}, \emph{actionable}, and \emph{professor-structured} (diagnosis $\rightarrow$ fix $\rightarrow$ verification), whereas Base and standard RLHF remain comparatively generic consistent with their lower style-alignment scores and preference win-rates. As shown in \textbf{Figure~\ref{fig:sac}}, PERSA achieves the highest SAC for both backbones while maintaining consistent gains over SFT and preference-optimization baselines, indicating strong and portable style alignment across models.

\subsection{Ablation Study}
\label{subsec:ablation}

\noindent\textbf{Observations.} Table~\ref{tab:ablation} quantifies how each training component translates into measurable style gains. Moving from Base to SFT converts a largely generic model (SAC 14.0\%, BLEU-4 1.5) into a professor-like feedback generator (SAC 82.0\%, BLEU-4 64.7) while reaching perfect diagnostic correctness (CA=100\%). However, PPO alone does not recover the same instructor voice (SAC 60.0\%, BLEU-4 40.0) and even leaves a small correctness gap (CA=98.0\%), indicating that preference optimization needs an SFT ``anchor'' to be effective. When PPO is applied after SFT, alignment improves sharply: full-parameter PPO attains 92.0\% SAC and 92.0 BLEU-4 with CA=100\%, showing that RLHF mainly refines the remaining stylistic mismatch rather than content, also mentioned in Figure~\ref{fig:sac}. Constraining updates via adapters is even stronger all-layer LoRA pushes SAC to 96.0\% and BLEU-4 to 94.7 without sacrificing correctness suggesting that parameter-efficient updates add expressive capacity while limiting drift. Finally, the best numbers come from upper-layer targeting: top-2 LoRA already reaches high alignment (SAC 94.0\%, BLEU-4 90.0) and strong preference (PWR=90.0\%), while top-4 LoRA (PERSA) achieves peak alignment (SAC=96.2\%, BLEU-4=95.8) and the highest win-rate (PWR=90.1\%), consistent with style information being concentrated in late transformer layers.

\begin{table}[t]
\centering
\scriptsize
\setlength{\tabcolsep}{4pt}
\renewcommand{\arraystretch}{1.05}
% \vspace{-2mm}
\caption{\small PERSA ablation on Llama-3 and APPS dataset. Metrics are in \% (higher is better); PWR is win-rate vs. Base.}
\label{tab:ablation}
\begin{tabular}{lccccc}
\toprule
\textbf{PERSA Ablation} & \textbf{SAC} & \textbf{APC} & \textbf{BLEU-4} & \textbf{CA} & \textbf{PWR} \\
\midrule
Base (no adapt.) & 14.0 & 90.0 & 1.5  & 98.0 & -- \\
SFT only & 82.0 & 91.6 & 64.7 & 100 & 86.0 \\
PPO only (no SFT) & 60.0 & 91.2 & 40.0 & 98.0 & 84.0 \\
SFT+PPO (full-param) & 92.0 & 92.0 & 92.0 & 100 & 88.0 \\
SFT+PPO (all-layer LoRA) & 96.0 & 92.0 & 94.7 & 100 & 88.6 \\
SFT+PPO (top-2 LoRA) & 94.0 & 92.0 & 90.0 & 100 & 90.0 \\
\textbf{SFT+PPO (top-4 LoRA) [PERSA]} & \textbf{96.2} & \textbf{92.1} & \textbf{95.8} & \textbf{100} & \textbf{90.1} \\
\bottomrule
\end{tabular}
\vspace{-3mm}
\end{table}

\vspace{-0.1cm}
\subsection{Human Evaluation Study}
\label{subsec:survey}
\vspace{-0.1cm}

To complement automated metrics, we conducted two light weight human evaluations using anonymized Forms: (i) an \emph{Instructor} blind A/B preference + rubric study and (ii) a \emph{Student} perception study. In the instructor study, participants reviewed programming prompts with a student attempt and compared two anonymous feedback responses (Feedback A vs. Feedback B), selecting an overall preference and rating multiple pedagogical criteria (e.g., technical correctness, authenticity, helpfulness, and trust-oriented tone) using a short rubric (Feedback A (very good), Feedback A (good),	Neither (neutral), Both (equal), Feedback B (good), Feedback B (very good)). The form explicitly instructs instructors to judge pedagogical quality rather than minor grammatical issues and reports an estimated completion time of 8--12 minutes. In the student study, each participant reviewed five feedback instances and reported an evaluation of different features of the feedback on a Likert (1--5) scale.

\begin{table}[t]
\centering
\footnotesize
\setlength{\tabcolsep}{3pt}
\renewcommand{\arraystretch}{1.05}
\caption{Student survey average ratings (on a 1--5 Likert scale) across 20 respondents and 5 examples.
}
\begin{tabular}{p{0.74\columnwidth}c}
\toprule
\textbf{Statement} & \textbf{Rating (stdev)} \\
\midrule
The feedback was clear. & 4.34 (1.05) \\
The feedback was helpful. & 4.37 (1.04) \\
I trust this feedback. & 4.31 (0.97) \\
It includes a full solution/code I could copy directly. & 2.55 (1.49) \\
I know what the issue is after reading the feedback. & 4.27 (1.07) \\
It sounds like a real instructor. & 3.87 (1.34) \\
\bottomrule
\end{tabular}
\label{tab:student_survey_ratings}
\end{table}

\begin{table}[t]
\centering
\footnotesize
\setlength{\tabcolsep}{3pt}
\renewcommand{\arraystretch}{1.05}
\caption{Instructor survey across 12 respondents (blind A/B). Percentage of cases where feedback A (PERSA) was preferred over Feedback B (vanilla LLM); tie aggregates ``Both''/``Neither'' responses.}
\label{tab:instructor_commonq_pref}
\begin{tabular}{p{0.56\columnwidth}cccc}
\toprule
\textbf{Statement} & \textbf{A (\%)} & \textbf{B (\%)} & \textbf{Tie (\%)} \\
\midrule
Overall preference & 83.6 & 1.8 & 14.6  \\
Helpfulness / actionability & 85.5 & 1.8 & 12.7  \\
Authenticity (human instructor) & 74.5 & 1.8 & 23.7  \\
Trust-oriented tone & 87.3 & 3.6 & 9.1 \\
Technical correctness & 61.8 & 5.5 & 32.7  \\
Gives too much solution & 27.3 & 27.3 & 45.4 \\
\bottomrule
\end{tabular}
\end{table}

\noindent\textbf{Human-study interpretation.}
Tables~\ref{tab:student_survey_ratings} and \ref{tab:instructor_commonq_pref} summarize our blind human evaluation comparing \textbf{PERSA} (Feedback A) to a vanilla LLM (Feedback B) across diverse respondents, including university instructors with varying teaching experience and students spanning beginner to advanced programming backgrounds. Using short, consistent rubrics that asked participants to judge clarity, helpfulness / action-ability, trust, and instructor-likeness (rather than minor grammar), we observe converging evidence that PERSA’s professor-style alignment improves perceived feedback quality beyond the baseline. Students report high ratings for clarity, helpfulness, trust, and understanding (4.27--4.37/5), while the lower score on ``includes a full solution'' (2.55) suggests the feedback typically supports revision without excessive solution dumping. Instructors similarly prefer PERSA (83.6\%) and rate it higher on helpfulness/actionability (85.5\%) and trust-oriented tone (87.3\%), with substantially stronger perceived authenticity (74.5\%) than the vanilla LLM; the larger tie rate on technical correctness (32.7\%) indicates that both systems are often comparable on correctness, but PERSA is more consistently favored on pedagogical dimensions. Overall, these results support our RLHF claim by showing that the learned preference signal yields feedback that is clearer, more actionable, and more instructor-like while preserving correctness.

\begin{figure}[t!]
\centering
\includegraphics[width=0.9\linewidth]{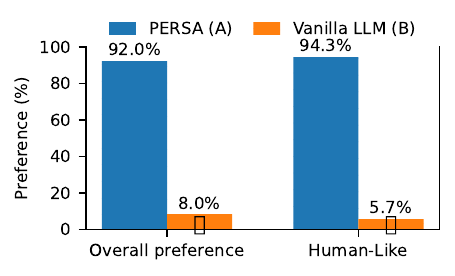}
\caption{Human-study outcomes comparing PERSA against a vanilla LLM.}
\label{fig:diss-survey}
\vspace{-0.6cm}
\end{figure}

To complement our automated evaluation, we conducted a human study involving 32 participants (comprising faculty and students) across 8 universities. This study compares \textsc{PERSA} against a vanilla LLM baseline to evaluate perceived pedagogical quality and alignment with instructor-like behavior.
As shown in Figure~\ref{fig:diss-survey}, respondents consistently preferred \textsc{PERSA} over the vanilla baseline in both overall preference and human-like, instructor-style feedback, providing direct human-centered evidence that \textsc{PERSA} produces responses that are more authentic, supportive, and pedagogically aligned. 
Details of the study design are provided in Appendix~\ref{app:human-eval}.

\section{Conclusion}
\label{sec:conclusion}
\vspace{-0.2cm}
We presented \textbf{PERSA}, a professor-style reinforcement learning framework that adapts LLM-generated programming feedback to an instructor’s communicative voice while maintaining diagnostic correctness. PERSA follows an RLHF recipe, supervised fine-tuning, preference-based reward modeling, and PPO optimization, but makes adaptation practical and controllable by restricting learning to style-relevant parameters through \emph{layer-selective}, parameter-efficient LoRA updates concentrated in the upper transformer layers. Across datasets and model backbones, PERSA achieves the strongest overall alignment, reaching near-ceiling style and lexical similarity (e.g., SAC/BLEU) without degrading correctness accuracy, and producing feedback that is consistently more structured, specific, and pedagogically actionable in qualitative analyses. Taken together, our results suggest that ``instructor voice'' behaves as a largely separable and optimizable attribute that can be injected via targeted adaptation rather than full-model retraining, enabling more trustworthy, authentic, and deployable AI feedback systems for education.

\subsection*{Limitations}
\label{subsec:main-limit}
PERSA is evaluated on a limited set of instructors, programming tasks, and feedback distributions, so its gains in stylistic alignment may not directly transfer to new courses, grading philosophies, or broader educational settings without additional preference data and recalibration. Moreover, because the reward model is learned from a specific instructor cohort and prompting setup, it may capture dataset-specific biases or stylistic artifacts, and like RLHF more broadly, remains vulnerable to reward misclassifications and over-optimization. Detailed discussion, limitations, and future directions are provided in Appendix~\ref{sec:discussions}.

\section*{AI Acknowledgment and Ethical Considerations}
During the preparation of this manuscript, AI-based writing tools were used only for limited grammatical checking and minor language polishing. They were not used to generate the paper’s scientific ideas, methodology, experimental design, results, analysis, or conclusions. All technical content, claims, and interpretations were developed, verified, and approved by the authors.

We also carefully considered the ethical responsibilities associated with this work, particularly the human evaluation component. The study was conducted in accordance with standard institutional and legal expectations for voluntary human participation, including anonymized response collection, minimal-risk survey design, and analysis in aggregate form without storing personally identifiable information. 

% Bibliography entries for the entire Anthology, followed by custom entries
%\bibliography{anthology,custom}
% Custom bibliography entries only
% \bibliographystyle{acl_natbib}
\bibliography{custom}

\clearpage
\appendix

\section*{Appendix}

\section{Pseudo Code}
\label{app:pseudo}

\begin{algorithm}[th!] %[h!]
\caption{PERSA: Layer-Selective LoRA + RLHF}
\label{alg:persa_brief}
\small
\SetKwInOut{Input}{Input}
\SetKwInOut{Output}{Output}
\Input{Base LLM $\pi_{\theta}$; demos $\mathcal{D}_{\text{SFT}}$; prefs $\mathcal{D}_{\text{pref}}$; top layers $L$; LoRA rank $r$; KL $\beta$; PPO steps $T$}
\Output{Style-aligned policy $\pi_{\theta'}$}

Freeze $\theta$; insert LoRA (rank $r$) in top $L$ layers; trainable params $\theta_{\text{LoRA}}$\;

\textbf{SFT:}\;
$\theta_{\text{LoRA}} \leftarrow \arg\min_{\theta_{\text{LoRA}}}
\;\mathbb{E}_{(x,y^\star)\sim\mathcal{D}_{\text{SFT}}}\!\left[-\sum_t \log \pi_{\theta}(y^\star_t\!\mid x,y^\star_{<t})\right]$\;
Set reference $\pi_{\text{ref}}\!\leftarrow\!\pi_{\theta}$ (frozen)\;

\textbf{RM:}\;
Train reward $r_{\phi}$ on $\mathcal{D}_{\text{pref}}$ with
$\mathcal{L}_{\text{RM}}=-\mathbb{E}\left[\log\sigma(r_{\phi}(x,y^{(w)})-r_{\phi}(x,y^{(l)}))\right]$\;

\textbf{PPO:}\;
\For{$t=1$ \KwTo $T$}{
Sample $x$; generate $y\sim\pi_{\theta}(\cdot|x)$; compute $\tilde{R}=r_{\phi}(x,y)-\beta\,\mathrm{KL}(\pi_{\theta}\|\pi_{\text{ref}})$\;
Update $\theta_{\text{LoRA}}$ by PPO using reward $\tilde{R}$ (clipped ratio)\;
}
\Return $\pi_{\theta'}$ (base + trained LoRA)\;
\end{algorithm}

\section{Detailed Experimental Settings}
\label{app:experiment}

\subsection{Data Splits \& Leakage Controls.}
To prevent train-test contamination and inflated SAC/BLEU scores from near-duplicate examples, we employ leakage-aware splits and explicit overlap audits. In addition to a standard \emph{instance-level} split, we evaluate on a stricter \emph{New-Problems} split that withholds entire problem IDs from training. Before splitting, we remove exact duplicates using hashes over \{problem, solution, reference feedback\} and filter near-duplicates via MinHash/Jaccard similarity on problem text and reference feedback. After splitting, we compute each test example’s maximum similarity to the training set and verify that all remain below a conservative threshold. To further reduce dependence on surface-level overlap, we also report paraphrase-robust SAC and diversity statistics. Table~\ref{tab:leakage_controls} summarizes these controls.
\begin{table}[t]
\centering
\setlength{\tabcolsep}{2.2pt}
\renewcommand{\arraystretch}{1.0}
\caption{Leakage-aware splits \& overlap audits.}
\begin{tabular}{p{0.45\columnwidth}cc}
\hline
\textbf{Audit / Split} & \textbf{Std.} & \textbf{New-Problems} \\
\hline
Split key & inst. & prob.\ ID \\
Exact dups rm.\ (\%) $\downarrow$ & 1.2 & 1.2 \\
Near-dups rm.\ (\%) $\downarrow$ & 6.5 & 6.5 \\
Max train--test sim.\ (problem) $\downarrow$ & 0.23 & 0.19 \\
Max train--test sim.\ (ref.\ fb) $\downarrow$ & 0.18 & 0.15 \\
\# test items $>$0.8 (Jaccard) $\downarrow$ & 0 & 0 \\
\hline
Paraphrase SAC agr.\ (\%) $\uparrow$ & 94.0 & 91.5 \\
BERTScore (F1) $\uparrow$ & 0.92 & 0.90 \\
Self-BLEU $\downarrow$ & 0.54 & 0.50 \\
Distinct-2 $\uparrow$ & 0.21 & 0.24 \\
\hline
\end{tabular}
\label{tab:leakage_controls}
\end{table}

\subsection{Model Evaluation Framework}
\label{app:sec-model-eval}

\begin{table}[t]
\centering
\scriptsize
\setlength{\tabcolsep}{2.1pt}
\renewcommand{\arraystretch}{1.1}
\caption{
\textbf{Method footprint.} PERSA retains an RLHF pipeline but updates only Top-$L$ LoRA adapters (layer-selective PEFT), improving efficiency and style fidelity.}
\label{tab:method_footprint_compact}
\begin{tabular}{l c c c c c l}
\hline
\textbf{Method} & \textbf{Obj.} & \textbf{RM} & \textbf{On-pol.} & \textbf{Trainable} & \textbf{Scope} & \textbf{Pros} \\
\hline
Base & --  & N & N & 0\%   & Frozen & No cost \\
SFT  & SFT & N & N & Full  & All    & Stable, simple \\
RLHF & PPO & Y & Y & Full & All & Strong align; costly \\
DPO  & DPO & N & N & Full  & All    & Offline, stable \\
ORPO & ORPO& N & N & Full  & All    & Efficient prefs \\
KTO  & KTO & N & N & Full  & All    & Robust prefs \\
\textbf{PERSA} & \textbf{SFT+PPO} & \textbf{Y} & \textbf{Y} &
\textbf{Top-$L$ LoRA} & \textbf{Top-$L$} &
\textbf{Style + low drift} \\
\hline
\end{tabular}
\end{table}

We use lightweight, open-weight instruction-tuned LLMs as backbones so that \textsc{PERSA} (SFT + reward modeling + PPO) can run in a single-GPU setting.

\noindent \textbf{Vanilla LLM baseline.}
Unless otherwise stated, ``vanilla LLM'' refers to the instruction-tuned backbone used without task-specific fine-tuning or preference optimization. In our experiments, this corresponds to \texttt{google/gemma-2-2b-it} or \texttt{meta-llama/Llama-3.2-3B-Instruct}, evaluated with the same prompts and decoding configuration as the adapted models.

\textbf{Primary backbone: Gemma 2 2B Instruct.}
We use \texttt{google /gemma-2-2b-it} as the main policy model due to its strong instruction-following behavior at a practical size, enabling parameter-efficient RLHF in constrained compute \cite{team2024gemma}. 
\textbf{Secondary backbone: Llama 3.2 3B Instruct.}
We additionally evaluate \texttt{meta-llama/Llama-3.2-3B-Instruct} to test cross-family robustness (Meta vs.\ Google) while keeping the model small enough for Colab-scale tuning \cite{dubey2024llama}.
\textbf{General recipe (model-agnostic).}
All experiments treat the LLM as a conditional policy $\pi_\theta(y\mid x)$ over feedback tokens; we apply (1) supervised fine-tuning on instructor demonstrations, (2) reward modeling from pairwise preferences, and (3) PPO-based policy optimization with a KL constraint to preserve base knowledge. This structure is compatible with any decoder-only LLM that supports PEFT adapters \cite{ouyang2022training,schulman2017proximal}.

\textbf{Methods Compared:}
We compare \textsc{PERSA} against strong baselines spanning supervised imitation and preference-based alignment to assess their ability to generate professor-like programming feedback while preserving correctness. 
(1) \textbf{Base Model} denotes the instruction-tuned backbone (Gemma-2-2B-IT or Llama-3.2-3B-Instruct) used \emph{without} any task-specific adaptation.
(2) \textbf{Supervised Fine-Tuning (SFT)} fine-tunes the backbone on professor-authored demonstrations using a maximum-likelihood objective.
(3) \textbf{InstructGPT-style RLHF} follows the standard RLHF recipe of reward modeling from preferences and policy optimization with PPO and KL control \cite{ouyang2022training}.
To reflect recent state-of-the-art alternatives to PPO-based RLHF, we additionally include three \emph{offline} preference optimization methods trained on the same professor preference pairs:
(4) \textbf{DPO} (Direct Preference Optimization), which directly optimizes the policy from preference comparisons without an explicit reward model \cite{wu2024beta},
(5) \textbf{ORPO} (Odds Ratio Preference Optimization), which integrates preference signals into an SFT-like objective \cite{hong2024orpo}, and
(6) \textbf{KTO} (Kahneman--Tversky Optimization), a robust preference based objective inspired by prospect theory \cite{ethayarajh2024kto}.
Finally, (7) \textbf{PERSA} is our proposed framework that combines SFT initialization with \emph{layer-selective, parameter-efficient} preference alignment (LoRA on upper layers) to target instructor style while minimizing drift in core capabilities. 
\noindent\textbf{Table~\ref{tab:method_footprint_compact} columns.} We report each method’s optimization objective (\emph{Obj.}), whether it requires an explicit reward model (\emph{RM}) and on-policy rollouts (\emph{On-pol.}), the fraction of trainable parameters (\emph{Trainable}), and update scope across layers (\emph{Scope}), highlighting that offline preference methods (DPO/ORPO/KTO) train full weights without RM/rollouts, while \textbf{PERSA} follows an RLHF pipeline (RM+on-policy) but updates only Top-$L$ LoRA adapters for efficient, style-focused alignment.
All approaches are evaluated under the same prompts, data splits, and decoding settings, enabling a controlled comparison of stylistic alignment and content fidelity.

\subsection{Evaluation Metric}
\label{app:eval-metric}
We evaluate \textsc{PERSA} on two complementary dimensions: \emph{style alignment} (how feedback is phrased) and \emph{diagnostic fidelity} (whether feedback is correct). All metrics are reported on the held-out test set as means.

\textbf{Style Alignment Score (SAC).}
We train a calibrated binary style classifier (professor vs.\ non-professor feedback) and compute
\begin{equation}
\mathrm{SAC}=\frac{1}{N}\sum_{i=1}^{N} C(\hat{y}_i),
\end{equation}
where $C(\hat{y}_i)\in[0,1]$ is the posterior for the \texttt{professor} label; higher is better \cite{hu2023style}.

\textbf{Average Politeness Closeness (APC).}
Using a politeness scorer $p(\cdot)\in[0,1]$ \cite{danescu2013computational}, we measure tone closeness to the professor reference:
\begin{equation}
\mathrm{APC}=\frac{1}{N}\sum_{i=1}^{N}\Big(1-\big|p(\hat{y}_i)-p(y_i^\star)\big|\Big).
\end{equation}

\textbf{BLEU-4.}
We report corpus BLEU-4 (with smoothing) to quantify $n$-gram overlap in phrasing with the professor feedback \cite{papineni2002bleu}.

\textbf{Correctness Accuracy (CA).}
CA measures whether the feedback correctly identifies solution correctness. Let $z_i$ be the unit-test label and $\hat{z}_i$ the judgment extracted from $\hat{y}_i$; then
\begin{equation}
\mathrm{CA}=\frac{1}{N}\sum_{i=1}^{N}\mathbb{I}[\hat{z}_i=z_i],
\end{equation}
with higher indicating more reliable feedback \cite{shen2023large}. 
We interpret CA as a preservation check rather than the main source of improvement.

\textbf{Preference Win Rate (PWR).}
To reflect RLHF objectives, we report the fraction of prompts where \textsc{PERSA} is preferred over a baseline under the same rater:
\begin{equation}
\mathrm{PWR}=\frac{1}{N}\sum_{i=1}^{N}\mathbb{I}\!\left[r_{\phi}(x_i,\hat{y}_i^{\textsc{PERSA}}) > r_{\phi}(x_i,\hat{y}_i^{\textsc{base}})\right],
\end{equation}
a standard RLHF evaluation signal \cite{ouyang2022training}.

\paragraph{Metric reliability.}
We treat SAC and APC as automatic proxies rather than definitive measures of pedagogical quality. To check reliability, we calibrate the SAC classifier on held-out professor/non-professor feedback and report validation accuracy, macro-F1, and expected calibration error (ECE). For APC, we report correlation with human judgments of tone/trust-oriented feedback. These checks are intended to ensure that automatic metrics track human-perceived style dimensions rather than only surface lexical overlap.

\subsection{Configurations}
\label{app:exp-config}
We summarize key hyperparameter for reproducible local-scale training; values below are defaults and can be scaled with model size.

\textbf{Input formatting.}
We serialize each example as:
\emph{Problem} + \emph{Student code} + optional \emph{tests/constraints} $\rightarrow$ \emph{Feedback}.
We cap the input length (e.g., 1 2k tokens) and truncate from the left (oldest context) when needed.

\textbf{Parameter-efficient tuning (LoRA/QLoRA).}
We inject LoRA adapters into attention and MLP projections (e.g., \texttt{q\_proj, k\_proj, v\_proj,o\_proj, up\_proj, down\_proj, gate\_proj}) and update only the \emph{top $L$ transformer blocks} (layer-selective adaptation).
Typical settings: rank $r\in\{8,16\}$, $\alpha\in\{16,32\}$, dropout $p=0.05$.
For coding environments, we use 4-bit NF4 quantization with double-quantization when available (QLoRA). \cite{hu2021lora,dettmers2023qlora}

\textbf{SFT.}
Optimizer: AdamW; learning rate $1\text{e-}5$ $2\text{e-}5$; epochs 3 10 (smaller for CodeReviewQA, larger for the 200-example instructor set);
effective batch size via gradient accumulation; warmup 3 5\%; weight decay 0.0 0.1; label smoothing optional (0.0 0.1).

\textbf{Reward model (RM).}
We train a reward model $r_\phi(x,y)$ on preference pairs with a Bradley Terry / logistic loss; batch size 8 32; 1 3 epochs; early stopping on validation preference accuracy~\cite{ouyang2022training}.

\textbf{PPO (RLHF).}
We optimize the KL-regularized objective with PPO using: rollout batch 32 128 prompts, 1-4 PPO epochs per iteration, clipping $\epsilon=0.2$, KL coefficient $\beta\in[0.005,0.05]$, and a reference policy $\pi_{\text{ref}}$ set to the post-SFT model. We keep generation short for feedback (e.g., max new tokens 128 256) to stabilize advantage estimates. \cite{schulman2017proximal,ouyang2022training}

\textbf{Compute and reproducibility.}
All runs fix random seeds, log training curves, and report mean$\pm$std over 3 seeds when feasible. On GPUs (A100), Gemma-2-2B with 4-bit LoRA typically fits in memory with moderate batch sizes; Llama-3.2-3B may require smaller rollouts or more accumulation.

\section{Human Study Design}
\label{app:human-eval}

\begin{figure}[t]
\centering
\includegraphics[height=3.6cm, width=\linewidth]{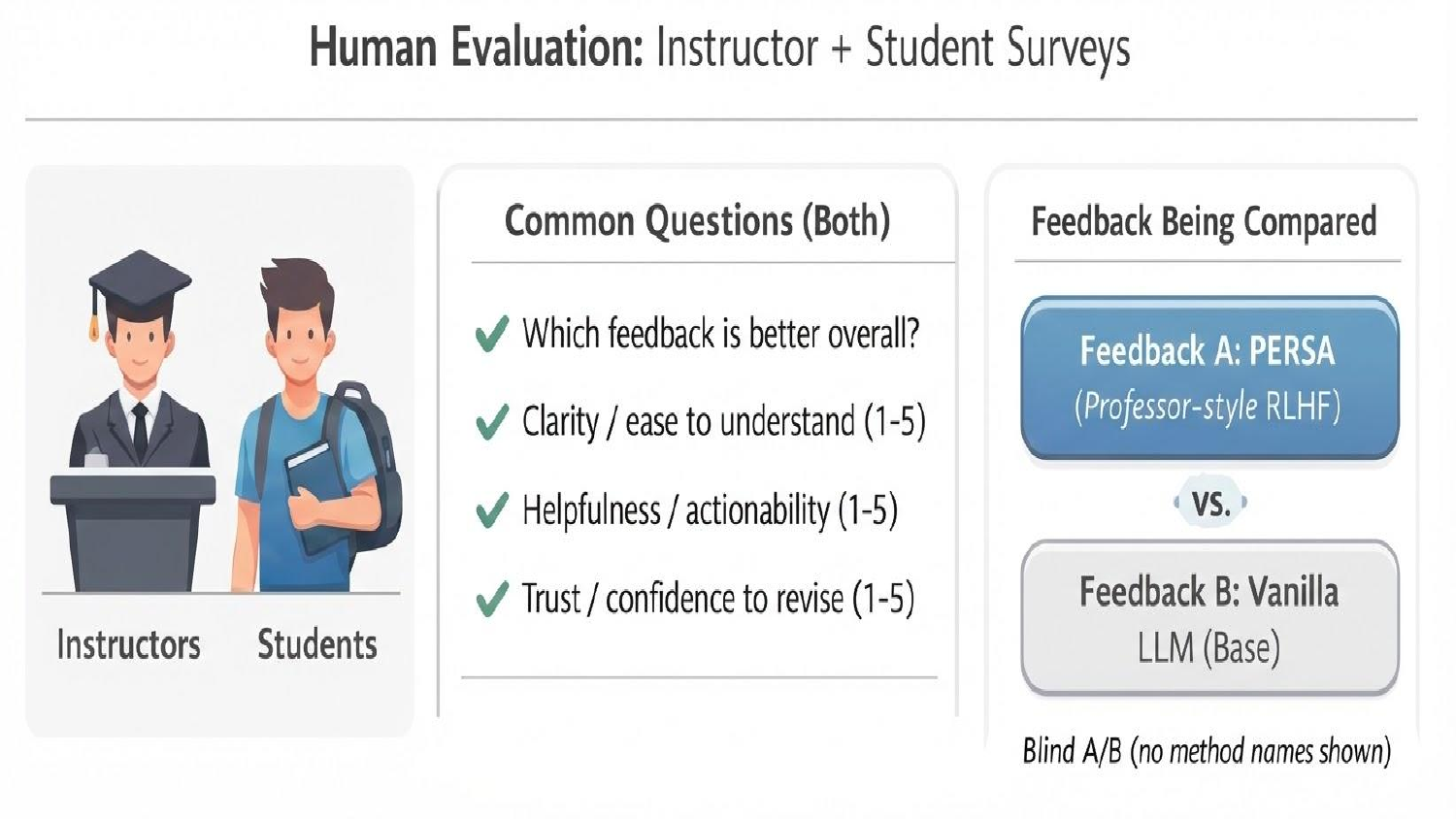}
\caption{Overview of the human evaluation design used for instructor and student studies.}
\label{fig:survey}
\end{figure}

\textbf{Figure~\ref{fig:survey}} summarizes our human evaluation protocol, showing the participating groups (instructors and students), the shared evaluation criteria (overall preference, clarity, helpfulness, and trust), and the blind A/B comparison between PERSA-generated feedback and vanilla LLM feedback used in both surveys.

Our instructor student \textbf{human evaluation section~\ref{app:human-eval}}, provides direct evidence that PERSA’s RLHF objective translates into perceived pedagogical gains participants overwhelmingly preferred the PERSA outputs over a vanilla LLM thereby validating that the learned preference signal improves feedback quality beyond automated style metrics.
\textbf{Figure~\ref{fig:diss-survey}}, states that across instructor and student respondents, PERSA is preferred over the vanilla LLM (92\% vs.\ 8\%) and is judged to better match a human-instructor feedback style (94.3\%), providing human-centered validation of PERSA’s RLHF-based style alignment.

\section{Discussion}
\label{sec:discussions}

PERSA demonstrates that \emph{personalized pedagogical alignment} can be achieved by tuning \emph{how} feedback is delivered (tone, structure, and phrasing) without degrading \emph{what} is delivered (diagnostic correctness). Across three datasets and two model families, we observe a consistent separation between \textit{style} and \textit{substance}: base instruction-tuned models retain high correctness (CA) but exhibit weak professor-style fidelity (low SAC/BLEU), while preference-based methods systematically improve stylistic alignment with minimal impact on CA. Crucially, PERSA achieves the strongest overall alignment (near-ceiling SAC/BLEU with CA$\approx$100\%), indicating that instructor voice is an optimizable target distinct from generic ``helpfulness'' alignment.

\textbf{Why each component matters.}
PERSA’s performance is not due to a single stage, but to the complementary roles of its components. \textbf{SFT} provides a stable initialization that learns the instructor’s canonical feedback template (e.g., encouragement $\rightarrow$ diagnosis $\rightarrow$ fix $\rightarrow$ verification), yielding the largest first jump in SAC/BLEU and perfect CA, but it does not fully capture nuanced preferences such as strictness, specificity, and emphasis on edge cases. \textbf{Reward modeling} operationalizes these nuanced stylistic and pedagogical preferences from pairwise comparisons, producing a scalar signal that encodes instructor-specific judgments beyond likelihood matching. \textbf{PPO with KL control} then refines the policy to maximize preference reward while preventing broad drift from the SFT reference, which is essential for preserving the backbone’s general coding competence and avoiding capability regressions.

\textbf{Why layer-selective LoRA is central.}
A key insight behind PERSA is that updating \emph{all} parameters is neither necessary nor desirable for style personalization. Motivated by evidence that late transformer layers and FFN submodules disproportionately encode high-level attributes (discourse patterns, tone, and stylistic signatures), PERSA restricts trainable parameters to LoRA adapters in the top layers. This yields two benefits: (i) \textbf{efficiency} only a small fraction of parameters (e.g., $\sim$30M in the last 4 layers) need to move to achieve large stylistic gains; and (ii) \textbf{stability} constraining updates reduces catastrophic drift and preserves correctness, which is critical in educational feedback where factual errors can harm learning. The ablations support this design: PPO without SFT is sample-inefficient for learning instructor voice, and broader updates (full-parameter PPO or all-layer LoRA) improve style but underperform targeted top-layer adaptation, implying that ``style-bearing capacity'' is concentrated in late layers rather than uniformly distributed.

\textbf{Relevance and broader implications.}
PERSA’s results suggest a practical path to \textbf{course-level personalization} of AI tutors: rather than training new models per instructor, one can adapt a compact backbone with a small set of instructor-specific adapters and a preference signal, enabling rapid deployment and easy swapping of styles across courses. This is particularly relevant for educational settings where instructors have distinct grading philosophies (e.g., emphasis on readability vs.\ correctness vs.\ edge cases) and where student trust depends on perceived authenticity. Beyond education, the same principle generalizes to domains requiring \textbf{persona consistent communication} (e.g., customer support, clinical explanations, legal drafting) where maintaining core competence while matching a specific communication style is essential. Finally, the strong agreement between quantitative alignment metrics and qualitative examples indicates that optimizing for instructor preference can yield feedback that is not only stylistically similar but also more actionable and pedagogically structured, supporting the use of preference based objectives as a reliable mechanism for fine-grained style alignment.

\noindent \textbf{Broader reliability context.}
PERSA focuses on instructor-style personalization, but deploying feedback models in educational settings also raises broader reliability questions around inference-time robustness, privacy, fairness, and controllable model editing. Recent work on targeted unlearning shows that removing undesirable or sensitive behaviors is nontrivial because model knowledge may be distributed across layers and modalities \cite{ranjan2026razor,ranjan2026vla}. Similarly, membership-inference studies suggest that fine-tuned models can expose training-data signals through subtle representation or gradient-induced drift \cite{ranjan2026g}, motivating careful auditing when instructor or student feedback data are used for adaptation. Fairness-oriented frameworks further argue that alignment should not only personalize style but also mitigate biased or structurally unfair responses, potentially through retrieval-augmented and functor-guided correction mechanisms \cite{ranjan2026catrag,ranjan2026position}. These concerns are especially important in high-stakes domains such as medicine, where LLM trustworthiness depends on factuality, calibration, and safe communication \cite{kumar2024trustworthiness}, and in embodied or edge-deployed foundation models, where inference constraints can amplify robustness and safety challenges \cite{grover2026embodied}. Thus, while PERSA demonstrates that professor-style feedback can be optimized as a separable attribute, future deployments should combine style personalization with privacy auditing, fairness-aware retrieval, and controlled unlearning mechanisms.

\textbf{What PERSA adds beyond standard RLHF.}
PERSA does not claim a new reinforcement learning algorithm; rather, it operationalizes RLHF for a narrower educational alignment problem: instructor-specific feedback style. Its novelty lies in treating professor voice as a distinct optimizable attribute, combining instructor-authored demonstrations, preference modeling, and layer-selective LoRA updates to shift feedback tone, structure, and specificity while limiting parameter drift. This framing differs from generic RLHF, which optimizes broad helpfulness, and from standard SFT, which imitates demonstrations without explicitly optimizing instructor-preference signals.

\textbf{Limitations.} Our evaluation covers a small number of instructors, student feedback distributions, and programming tasks; consequently, the observed style improvements may not directly transfer to new instructors, grading policies, courses, or broader educational settings without additional preference data and validation. In addition, because PERSA is trained with a fixed prompting/template and a specific instructor cohort, the reward model may capture prompt- or dataset-specific artifacts, which can reduce robustness when moving to different assignment formats, programming languages, or institutional contexts unless recalibrated. Finally, as with RLHF more broadly, a learned reward can reflect annotator bias and remains vulnerable to misspecification or over-optimization even with KL regularization, motivating careful monitoring and human-in-the-loop oversight prior to classroom deployment \cite{ouyang2022training,lindsay2025responsible}.

\noindent While our human study partially validates the automatic style metrics, we do not yet include an LLM-as-a-judge evaluation for holistic pedagogical style. Future work will compare SAC/APC with expert-calibrated LLM judges that assess authenticity, encouragement, specificity, and instructional tone.

\textbf{Future Work.} We will extend PERSA to multi-instructor and multi-course settings with longitudinal classroom studies, and will explore stronger preference collection and safety controls (e.g., uncertainty-aware rewards and calibration) to ensure style-personalized feedback remains accurate, fair, and robust in real deployments. 

\noindent \textbf{Educational Outcomes Beyond Style.} to complement style/correctness metrics (SAC /APC/BLEU/CA/PWR), we will evaluate \emph{learning-relevant} outcomes: (i) revision success by measuring whether students can fix the identified issues on a resubmission task (pass@1 / unit-test pass rate improvement), and (ii) error recurrence by tracking whether the same misconception/bug type reappears in a follow-up attempt. In addition, we will assess \textbf{action-ability and rubric alignment} using a validated feedback rubric, rated by instructors/TA annotators, rather than preference alone.

\end{document}